\documentclass[final,3p,times]{elsarticle}

\usepackage{graphicx}
\usepackage[export]{adjustbox}         
\usepackage{caption}
\usepackage{subcaption}
\usepackage{amssymb,amsmath}
\usepackage{booktabs}
\usepackage{multirow}
\usepackage{afterpage}
\usepackage{setspace}
\usepackage{breqn}
\usepackage{microtype}    
\usepackage{titlesec}     
\usepackage{fancyhdr}     
\usepackage[breaklinks,colorlinks=true,linkcolor=blue,citecolor=blue,urlcolor=blue]{hyperref}  
\usepackage[noabbrev,nameinlink]{cleveref}

\crefname{figure}{Fig.}{Figs.}  
\crefname{table}{Table}{Tables}  
\crefname{equation}{Eqn.}{Eqns.}  
\usepackage[font=small, labelfont=bf, skip=5pt]{caption}
\titlespacing*{\section}{0pt}{*1}{*0.5}
\journal{Materials Letters}

\begin{document}
\doublespacing
\begin{frontmatter}

\title{Enhancing Experimental Efficiency in Materials Design: A Comparative Study of Taguchi and Machine Learning Methods}

\author[a]{Shyam Prabhu\corref{cor}} \ead{shyamprabhu92@gmail.com, +91-7389382333}
\author[a]{P Akshay Kumar\corref{cor}}
\ead{akshaykumar1993@gmail.com}
\author[a]{Antov Selwinston}
\author[a]{Pavan Taduvai}
\author[a]{Shreya Bairi}
\author[a,b]{Rohit Batra\corref{cor}}
\ead{rbatra@iitm.ac.in, +91-4422574780}

\cortext[cor]{Corresponding authors}
\affiliation[a]{organization={Department of MME},
            addressline={IIT Madras}, 
            city={Chennai},
            postcode={600036}, 
            country={India}}
\affiliation[b]{organization={ CAMMD},
            addressline={IIT Madras}, 
            city={Chennai},
            postcode={600036}, 
            country={India}}

\begin{abstract}
Materials design problems often require optimizing multiple variables, rendering full factorial exploration impractical. full factorial exploration. Design of experiment (DOE) methods, such as Taguchi technique, are commonly used to efficiently sample the design space but they inherently lack the ability to capture non-linear dependency of process variables. In this work, we demonstrate how machine learning (ML) methods can be used to overcome these limitations. We compare the performance of Taguchi method against an active learning based Gaussian process regression (GPR) model in a wire arc additive manufacturing (WAAM) process to accurately predict aspects of bead geometry, including penetration depth, bead width, and height. While Taguchi method utilized a three-factor, five-level L25 orthogonal array to suggest weld parameters, the GPR model used an uncertainty-based exploration acquisition function coupled with latin hypercube sampling for initial training data. Accuracy and efficiency of both models was evaluated on 15 test cases, with GPR outperforming Taguchi in both metrics. This work applies to broader materials processing domain requiring efficient exploration of complex parameters.
\end{abstract}

\end{frontmatter}

\section{Introduction}\label{sec1}

Accurate process parameter selection is critical in wire arc additive manufacturing (WAAM), a layer-wise metal deposition process used across aerospace, automotive, and marine sectors. In WAAM process variables such as current, weld speed, and wire feed rate strongly influence bead geometry,  affecting both layer uniformity and part quality \cite{li2022}. The nonlinear dependence of weld quality on process parameters makes optimization for different electrode-substrate pairs experimentally intensive and costly \cite{luo2009}. Single weld bead trials precede multilayer builds to enable adaptive slicing and improve efficiency.

To reduce experimental load, statistical design of experiment (DOE) methods such as the Taguchi technique are commonly used \cite{montgomery2017}. By employing orthogonal arrays, Taguchi facilitates parameter screening with fewer trials. Prior studies have demonstrated its use in welding optimization---Dinovitzer et al. \cite{dinovitzer2019} used Taguchi L16 array in WAAM to find most important process variables and Lin et al. \cite{lin2014} combined Taguchi with grey relational analysis to optimize aluminum alloy welding. Though effective for trend analysis, Taguchi’s linear assumptions and lack of uncertainty quantification limit its predictive power in nonlinear, continuous design spaces.

\begin{figure*}[!htb]
\centering
\includegraphics[width=0.8\textwidth]{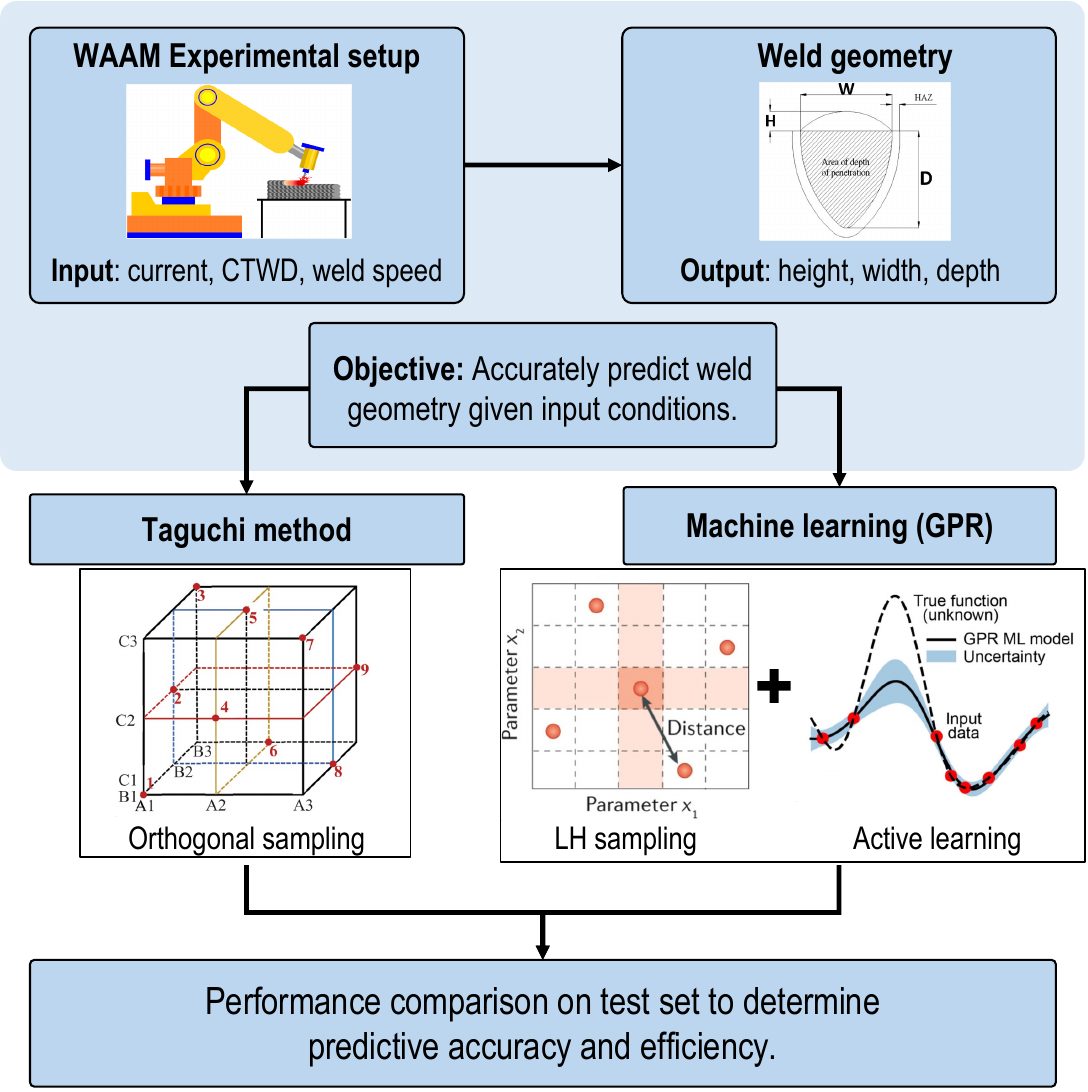}
\caption{Overview of the WAAM design problem comparing the accuracy and efficiency of both models. The Taguchi method uses orthogonal sampling \cite{montgomery2017}, while GPR-based method combines LHS \cite{batra2021} with uncertainty-based active learning \cite{deringer2021}. A test set of 15 experiments is used for evaluation.} 
\label{fig:1}
\end{figure*}

Machine learning (ML) techniques, such as neural networks (NN) and Gaussian process regression (GPR), offer improved modeling of such nonlinear interactions. In fact, NN-based approaches have been used to predict WAAM bead characteristics given wire feed speed and prior-layer geometry \cite{wang2021}, although such approach require extensive historical data to train an accurate model. To reduce these data requirements, Kim et al. \cite{kim2021} proposed an ANOVA-guided GPR framework, which achieved improved performance with fewer trials, though it remained limited by discretized input levels.

This study adopts a GPR framework with Latin hypercube sampling (LHS) \cite{iman2008} and active learning \cite{settles2009} to predict output bead geometry in the WAAM process, and compares its accuracy and efficiency against the Taguchi method (see Figure \ref{fig:1}). Taguchi applies a fixed L25 orthogonal array for parameter selection, while GPR begins with five LHS samples and iteratively selects new trials from high uncertainty regions using an active learning strategy \cite{batra2021}. Both models are evaluated on 15 independent test cases using root mean square error (RMSE) and coefficient of determination (R²) score \cite{geron2019}. GPR demonstrates greater flexibility in capturing nonlinear trends and achieves higher predictive accuracy with fewer experiments compared to the Taguchi method.

\begin{figure*}[!htb]
\centering
\includegraphics[width=0.8\textwidth]{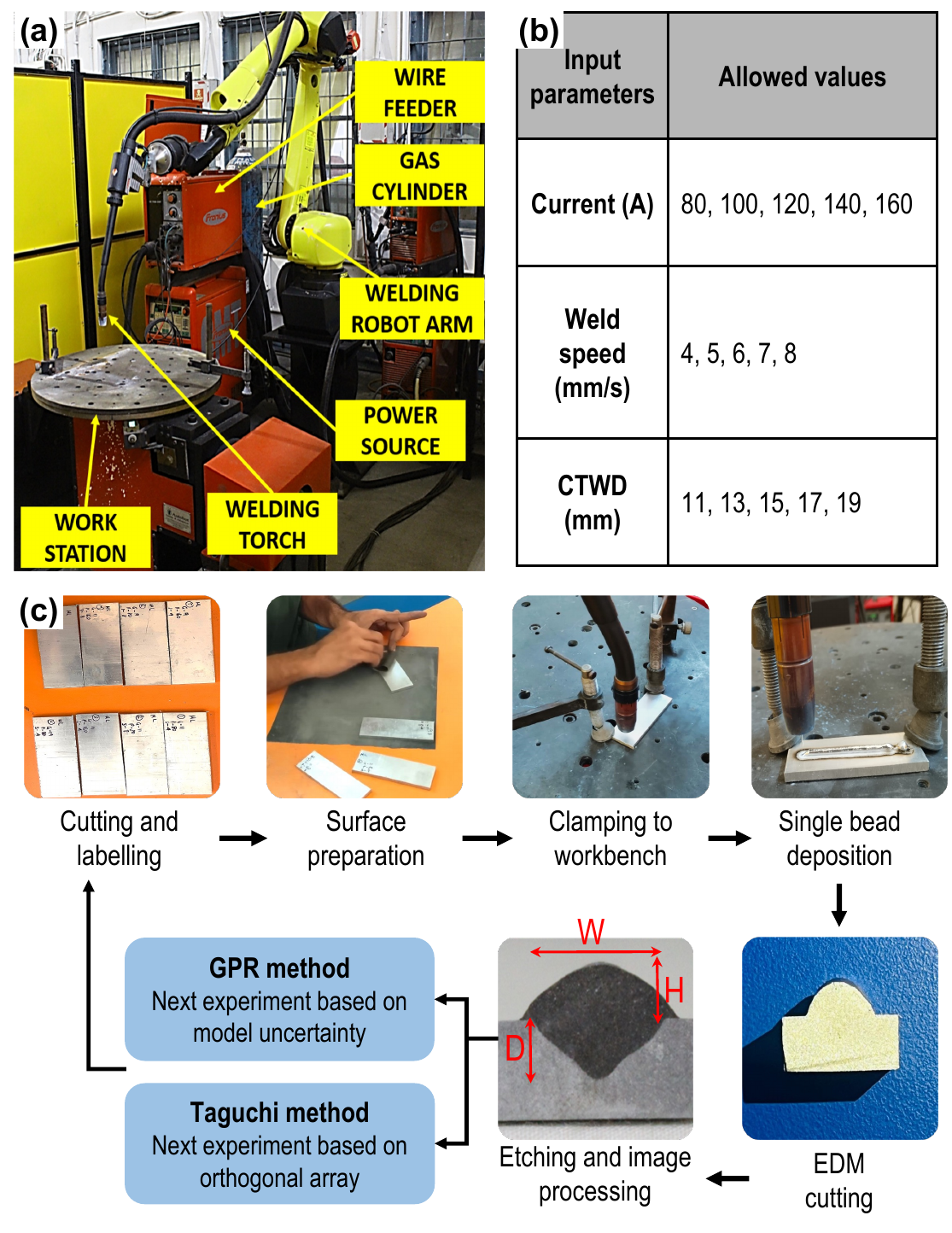}
\caption{(a) WAAM setup for single weld bead deposition. (b) Input design space. (c) Overall workflow involving sample preparation, deposition, measurement, and experiment selection for Taguchi and GPR models.}
\label{fig:2}
\end{figure*}

\section{Experimental procedure and methodology}\label{sec2}

Weld beads were deposited using Al6063 electrodes on Al4043 substrates using FANUC robotic system (see Figure~\ref{fig:2}(a)). A preset CMT C1658 program ensured synchronized current, voltage, and wire feed speed. Three process parameters---current, weld speed, and contact-tip-to-work-distance (CTWD)---were varied, while argon shielding gas was held constant at 15~L/min to maintain arc stability. Bead geometry, i.e. width ($W$), height ($H$), and penetration depth ($D$), was measured using \textit{ImageJ} software. To reduce dimensionality of the multi-objective problem, a single response parameter $Y$ was formulated as: $
Y = \log \left( \frac{ D \times 10^{10} + W \times 10^6 + H \times 10^2 }
{D + W + H} \right) \times 10
$. The formulation prioritizes penetration depth due to its role in preventing keyholing and balling \cite{mattera2024}. Overall, the design space comprised of 125 possible parameter combinations owing to the limitation imposed by the Taguchi L25 array. From this space, 15 combinations were selected as test cases to evaluate and compare the Taguchi and GPR models. RMSE and R² score were used to assess prediction accuracy \cite{geron2019}.

Current, weld speed, and CTWD were varied at five levels, yielding 
$5^3 = 125$ possible combinations (Figure~\ref{fig:2}(b)). The Taguchi method used an L25 orthogonal array to systematically select 25 representative design points. Figure~\ref{fig:2}(c) shows the Taguchi workflow.. Predictions for remaining 100 cases, including 15 test points, were made in Minitab using the main effects formula \cite{montgomery2017}:

\begin{equation}
\mu_{\text{predicted}} = \mu + \alpha_i + \beta_j + \gamma_k
\label{eq2}
\end{equation}

\noindent where $\mu$ is the overall mean response, and $\alpha_i$, $\beta_j$ and $\gamma_k$ represent the effect of current, weld speed, and  CTWD at levels $i$, $j$ and $k$, respectively. Predicted values were then compared with experimental results for the 15 test cases to assess model accuracy.

A GPR model was also developed to predict the response variable $Y$ using current, weld speed, and CTWD as inputs. Five LHS samples were used for initial training to ensure uniform coverage and reduce selection bias \cite{iman2008}. To improve accuracy with minimal experiments, active learning was used. The GPR model iteratively selected new data points from regions with the highest predictive uncertainty \cite{settles2009}. The model was then refined over 15 iterations (Figure~\ref{fig:2}(c)) with predictions on 15 test cases compared after each step against Taguchi method. A radial basis function kernel was employed to capture nonlinear relationships. Hyperparameters optimization, 5-fold cross-validation and feature normalization techniques improved model accuracy.

\section{Results and Discussion}\label{sec3}
\begin{figure*}[!htb]
\centering
\includegraphics[width=0.8\textwidth]{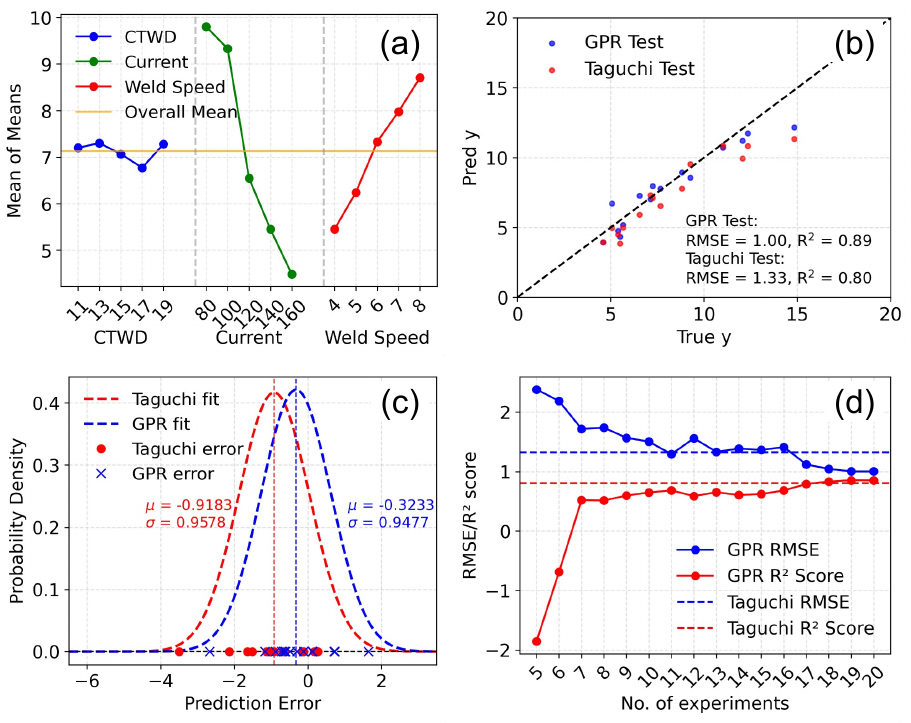}
\caption{(a) Mean-of means plot (b) Parity plot comparing GPR model and Taguchi method against experimental results (c) Error distribution of Taguchi and GPR models.(d) Evolution of RMSE and R$^2$ score of GPR model with each active learning iteration} 
\label{fig:3}
\end{figure*}

Figure~\ref{fig:3}(a) (mean-of-means plotof Taguchi method) shows that current and weld speed influenced response Y more than CTWD. Further, the predictive performance of both the Taguchi and GPR models, evaluated using the 15 test cases, is presented as parity plot in Figure~\ref{fig:3}(b). Taguchi model (with 25 training points) achieved a RMSE of 1.33 and an $R^2$ score of 0.80, indicating moderate accuracy and limited capability to model nonlinearity. The final GPR model (length scale = 2.68, noise level = 1.70) trained on just 20 points achieved an RMSE of 1.00 and an $R^2$ score of 0.89, outperforming the Taguchi method. Detailed model predictions and Python codes are provided in Supporting Information. As shown in Figure~\ref{fig:3}(c), the Taguchi predictions exhibit a systematic underprediction bias, whereas GPR errors are more symmetrically distributed around zero, highlighting another limitation of the Taguchi method likely due to the simplicity of the underlying model. Figure~\ref{fig:3}(d) shows progressive improvement in the accuracy of the GPR model as a function of increasing training data size. Notably, the GPR model approaches the performance of the Taguchi method (with 25 trials) after just seven experiments, and surpasses it after 17. Overall, the GPR+LHS approach demonstrated superior data efficiency and accuracy, requiring fewer experiments to exceed performance of Taguchi method.

\section{Conclusions}
This study compared the traditional Taguchi method with ML (GPR) active learning framework to efficiently predict WAAM weld bead geometry. The Taguchi method used a fixed L25 orthogonal array with 25 experiments, while GPR began with five LHS samples and iteratively selected new samples based on model uncertainty. GPR outperformed Taguchi in both accuracy and data efficiency, achieving lower RMSE and higher $R^2$ with just 17 experiments. Future work on the GPR framework can be extended to inverse design and expected improvement acquisition for optimizing weld parameters to achieve target bead geometry. Further, it could be applied to materials design and nonlinear process optimization, including powder metallurgy and additive manufacturing. 

\section{CRediT authorship contribution statement}

Shyam Prabhu: Conceptualization, Writing – original draft. 
P. Akshay Kumar: Methodology, Software. 
Antov Selwinston: Formal analysis, Visualization. 
Pavan Kumar Taduvai: Data curation. 
Shreya Bairi: Project administration. 
Rohit Batra: Supervision, Validation, Writing – review \& editing.

\section{Declaration of Competing Interest}

The authors declare no competing interests.

\section{Acknowledgement}
RB acknowledges support from CAMMD, IIT Madras, funded using IOE scheme (SP22231239CPETWOAMMHOC). We appreciate Dr. Murugaiyan Amirthalingam assistance in extending facilities of joining \& additive manufacturing (JAM) lab.

\appendix
\renewcommand{\refname}{Supplementary}

\renewcommand{\refname}{References}

\end{document}